\documentclass{article}
\usepackage[margin=1.5in]{geometry}

\usepackage{amsthm}
\usepackage{amsmath}
\usepackage{amsfonts}
\usepackage{bbm}
\usepackage{graphicx}

\usepackage{caption}
\usepackage{subcaption}

\usepackage{algorithm}
\usepackage{algorithmic}

\newcommand{\INDSTATE}[1][1]{\STATE\hspace{#1\algorithmicindent}}

\usepackage{thmtools, thm-restate} 

\newtheorem{theorem}{Theorem}
\newtheorem{lemma}{Lemma}

\newtheorem{assumption}{Assumption}

\usepackage[utf8]{inputenc} 
\usepackage{hyperref}       
\usepackage{url}            
\usepackage{booktabs}       
\usepackage{amsfonts}       
\usepackage{nicefrac}       
\usepackage{microtype}      
\usepackage{xcolor}         
\usepackage{natbib}

\newcommand{\Lim}[1]{\raisebox{0.5ex}{\scalebox{0.8}{$\displaystyle \lim_{#1}\;$}}}

\title{Replay For Safety}

\author{Liran Szlak, Ohad Shamir}

\begin{document}

\maketitle

\begin{abstract}
Experience replay \citep{lin1993reinforcement, mnih2015human} is a widely used technique to achieve efficient use of data and improved performance in RL algorithms. In experience replay, past transitions are stored in a memory buffer and re-used during learning. Various suggestions for sampling schemes from the replay buffer have been suggested in previous works, attempting to optimally choose those experiences which will most contribute to the convergence to an optimal policy.
Here, we give some conditions on the replay sampling scheme that will ensure convergence, focusing on the well-known Q-learning algorithm in the tabular setting. After establishing sufficient conditions for convergence, we turn to suggest a slightly different usage for experience replay - replaying memories in a biased manner as a means to change the properties of the resulting policy. We initiate a rigorous study of experience replay as a tool to control and modify the properties of the resulting policy. In particular, we show that using an appropriate biased sampling scheme can allow us to achieve a \emph{safe} policy. We believe that using experience replay as a biasing mechanism that allows controlling the resulting policy in desirable ways is an idea with promising potential for many applications. 
\end{abstract}

\section{Introduction}
In reinforcement learning, a learner interacts with the environment with the purpose of learning a policy that will maximize the long-term return- the expected sum of rewards obtained. A common practice in many off-policy reinforcement learning algorithms is to use experience replay \citep{lin1993reinforcement, mnih2015human}, where experiences collected by interacting with the environment are stored and re-used during learning. In experience replay, the learner is allowed to access previous experiences, and use them to update the decision making policy as if they were transitions currently sampled from the world. This allows for better sample efficiency as experiences are not only used once at the time of their occurrence, but many times during learning, which can be useful in situations where data acquirement is costly or difficult. Moreover, it has been suggested that experience replay can potentially improves performance by breaking the time and space correlation structure of experiences as they are sampled from the real world, allowing for policy updates not dependent on the current time and state and randomizing over the data.
A key question in experience replay is deciding which experiences to replay. Most works attempt to find a sampling scheme from the replay buffer that will contribute most to the progress of convergence to the optimal policy (e.g. prioritized replay \citep{schaul2015prioritized}, episodic-backward update \citep{lee2018sample}, etc). Some use experience replay as a mechanism for learning additional goals (e.g. Hindsight Experience Replay \citep{andrychowicz2017hindsight}).
Here, we suggest a different usage of experience replay - we notice that by choosing which experiences to replay, one can effectively reshape the distribution of rewards and next-states as seen by the learner, and thus affect the resulting policy. We suggest to utilize this property to lead the learner to select actions that ensure the resulting policy has some desirable properties. In particular, we suggest to use a biased sampling scheme that will lead the policy to be \emph{safe}.
Importantly, such observations are also found in human learning, where biases in the replay mechanism during learning has been observed in many studies, relating to the salience/arousal of the stimulus \citep{gruber2016post}, to fear-conditioining and to high-reward \citep{de2016awake}.  

As a first step, we suggest a framework for Q-learning with experience replay where experiences are sampled from the replay buffer using a sequence of replay probability functions, $\{w_t\}_{t=0}^\infty$, where $t$ denotes the $t^{th}$ time step in the learning process. We give sufficient conditions on the sequence of replay probability functions to ensure convergence of the algorithm, and characterize the convergence point as a function of the replay sampling scheme. 
We then turn to show how using a sampling scheme which biases experiences in a well-designed way, can give rise to desirable properties of the learned policy, and demonstrate this on the example of safety. 

In safe RL, we are concerned not only about achieving the maximal expected return, but also about respecting additional safety constraints, such as having low variance of the return \citep{sato2001td}, avoiding error states \citep{geibel2005risk}, or ensuring the worst-case performance of the policy \citep{heger1994consideration, gaskett2003reinforcement, tamar2013scaling}. The safety of the policy is important because due to the stochasticity of the environment, even the optimal policy may lead to undesirable outcomes that might occur with low probability.
Approaches to incorporate the notion of safety into RL can roughly be divided to modifications to the optimization criteria, and approaches that enforce some limitations on the exploration process. The latter include inserting prior or expert knowledge of the domain \citep{song2012efficient, de2009learning, quintia2013learning}, using demonstrations \citep{abbeel2010autonomous, tang2010parameterized}, or enforcing other limitations on the exploration process. The former include using a weighted sum of the expected return and a safety factor such as the variance of the return \citep{sato2001td, geibel2005risk}, using an exponential utility function \citep{howard1972risk, basu2008learning}, using a worst-case optimization criterion \citep{heger1994consideration, gaskett2003reinforcement, tamar2013scaling}, or using constraint optimization where safety constraints such as the probability to reach error-states, a lower bound on the return or other constraints are imposed \citep{moldovan2012safe, moldovan2012risk}.

Here, we show how using experience replay, one can learn a safe policy, where no change to the optimization criteria is performed, and no external knowledge or expert advice is required. We propose a novel replay scheme where experiences are replayed with probability proportional to the variance of the reward received in each state-action pair, and where in highly variable states the experiences that are replayed are the most negative ones seen thus far. This biases the effective reward and transition distribution, signaling the learner to avoid such highly variable sate-action pairs and results in a safer policy. As previously mentioned, the notion proposed in this work is not limited to safety or risk seeking behaviors alone, but can be used for various demands and desired policy properties, as we discuss later.

Finally, we show how our suggested sampling scheme is able to learn a safe policy in two grid world environments.  
We believe that using experience replay as a biasing mechanism, explicitly biasing the transition and reward distribution, offers a promising new usage of experience replay.
\section{Previous work}
\paragraph{Experience replay} \mbox{ } \\
Experience replay can be incorporated into various RL algorithms, and various methods for sampling from the memory buffer have been suggested and experimentally explored in different tasks. Some of these sampling schemes include prioritized experience replay \citep{schaul2015prioritized, wang2016sample}, where experiences which are important (in the sense of their TD error magnitude) are prioritized over others and sampled with increased rate, highly improving performance on the Atari benchmark, uniform sampling \citep{liu2017effects} where the memory buffer size is the parameter of interest, and a window size too small/large is shown to decrease performance, recency-based replay \citep{wang2019boosting} where recent experiences have higher probability to be chosen, emphasizing recent experiences over older ones, so that experiences collected with behavior policies far from the current one are less likely to be chosen.
\citet{andrychowicz2017hindsight} propose Hindsight Experience Replay, where not only real transitions are stored but also 'imagined' transitions to possible 'subgoals' which were not the goal of the learned task, allowing for higher sample efficiency and better learning in sparse reward domains. 
\citet{zhang2017deeper} experimentally demonstrate for Q-learning that combining online experience with replayed experience in every iteration allows for faster convergence and a stable learning curve under different memory buffer sizes.
Many others have proposed various sampling schemes from the replay buffer, in an attempt to maximally exploit the collected experiences in order to improve the learning rate and resulting policy. 
However, not many works have theoretically analyzed for which replay schemes can we expect convergence to take place. An exception is an interesting theoretical work by \citet{vanseijen2015deeper}that formulates a type of experience replay as a planning step as in the famous Dyna algorithm. A general framework of sampling schemes that ensure convergence is still lacking, and here we provide with a sufficient condition for convergence. In addition, not many works have used the replay mechanism for purposed other than improving learning rate or policy performance, and here we propose an intuitive and straightforward way to use experience replay for explicitly modifying the policy properties, in particular, its safety.
To the best of our knowledge, this is the first work explicitly biasing the sampling of experiences in the purpose of of manipulating the properties of the learned policy, rather than to achieve better performance.
\paragraph{Safe reinforcement learning} \label{sec:prev_safe_RL} \mbox{ } \\
The concept of Safety has taken many forms in the literature. \citet{heger1994consideration, gaskett2003reinforcement, tamar2013scaling} consider the worst-case criteria and try to learn a policy that maximizes the return $R = \sum_{t=0}^\infty \gamma^t r_t$, the sum of discounted future rewards, in the worst-case scenario. This is formulated as a minimax problem: $\max_{\pi \in \Pi} \min_{\omega \in \Omega^\pi} E_{\pi,\omega}[R]$, where $\Omega^\pi$ is a set of trajectories that occurs under policy $\pi$. The worst-case criteria was shown to be overly pessimistic in some situations \citep{gaskett2003reinforcement}, which can be more injurious than beneficial. In addition, this approach allows learning only risk-averse policies, whereas our approach allows learning risk-averse, risk-neutral, and risk seeking policies. Another line of works uses exponential utility functions \citep{howard1972risk, basu2008learning} where the return $R$ is transformed so that the objective criteria is: $\max_{\pi \in \Pi} \beta^{-1} \log(E_{\pi}[\exp(\beta R)])$. A Taylor expansion of this gives: $\max_{\pi \in \Pi} E_{\pi}[R] + \frac{\beta}{2} Var(R) + \mathcal{O}(\beta^2)$, indicating that setting $\beta<0$ gives a risk-averse objective, $\beta=0$ a risk neutral, and $\beta>0$ a risk seeking objective. The majority of works using the exponential utility criteria assume knowledge of the reward and transition probabilities, and do not handle non-deterministic reward functions, which is irrelevant to our setting. Other approaches optimize a weighted sum of the expected reward and a risk factor, such as the variance of the return \citep{sato2001td}. However, maximizing over this criteria has several drawbacks - first, it has been shown to be NP-hard. Second, it could lead to counter-intuitive policies \citep{mannor2011mean} (i.e. a policy that obtained exceptionally high rewards initially might seek to obtain low rewards later in order to keep the variance of the entire return low). Our method allows looking at variance from the point of view of single actions and choose actions that have low variance in their immediate reward, which bypasses this issue. Additionally, penalizing for high variance of the return treats variance that stems from positive and negative feedback the same, which can be avoided in our framework by carefully designing the replay scheme. The work most similar to ours is that of \citet{mihatsch2002risk}. Similarly to us, they do not directly change the optimization criteria, rather, they modify the TD error in a biased manner, by treating negative and positive TD-errors asymmetrically. By doing so, they bias the resulting policy. In contrary to their work, we do not explicitly change the TD-error, but bias the effective distribution of rewards and transitions by biasing the sampling from the replay buffer. \citet{mihatsch2002risk} show that their approach effectively allows interpolating between the standard maximization criteria of the expected return, and the worst-case criteria discussed earlier. Our approach also allows interpolating between the two considerations - risk and performance, by controlling the amount of replay performed. However, using biased replay allows treating different state-action pairs differently, which is not possible through biasing the TD-error. This allows creating a wider range of biases in the transition and reward distributions, such as biases that only effect those state-actions pairs that have very high variance in their immediate rewards, but do not effect others. 
Experience replay draws inspiration from a biological phenomenon - in animals and humans, replay events occur following a learning session \citep{momennejad2017offline, schapiro2018human, gruber2016post, de2016awake}. In some works, it has been shown that experiences are not uniformly replayed, and that fear-conditioining and high-reward both leads to more neuronal replay \citep{de2016awake}. Thus, biased experience replay can be a possible explanation to natural phenomena in humans and animals, such as risk-aversion. 
\section{Setting and algorithm}
\subsection{Setting} 
A reinforcement learning problem can be formulated as a Markov decision process (MDP). An MDP is a tuple $<S,A,P,R>$, where $S$ is a set of states, $A$ is a set of actions, $P$ is a transition probability that gives the probability to transition between states given an action: $\forall s,s',a : p(s'|s,a)$. For all state-action-state triplets, $r(s,a,s')$ is a bounded reward random variable with $E[r(s,a,s')] = R(s,a,s')$, $\forall s,a,s': r(s,a,s') \in [r_{min} r_{max}]$. Rewards are sampled i.i.d over all rounds per state-action pair.
At every time step $t = 0,1,2,...$, the learning agent perceives the current state of the environment $s_t$, and chooses an action $a_t$ from the set of possible actions at that state. The environment then stochastically moves to a new state $s_{t+1}$ according to the transition probability $P$, and sends a numerical reward signal $r_t(s_t, a_t,s_{t+1})$. The transition probability satisfies the Markov property, meaning, given the current state $s_t$ and chosen action $a_t$, the transition probability to a new state $s_{t+1}$ does not depend on previous states and actions.
In the basic scenario, the learner's objective is to learn a policy $\pi$, that maps state $s \in S$ to action $a \in A$, such that it maximizes the return- the expected discounted future sum of rewards from each state $s$ - $V^{\pi} \left( s \right) = \mathbb{E} \left[ \sum_{i=0}^\infty \gamma^i \cdot r_i | s_0,  \pi \right]$, where $\gamma \in (0,1)$ is a discount factor which discounts the value of future rewards.
 
\subsection{Q-Learning with convergent replay}\label{sec:Q_learning_with_convegrnt_replay}
We focus on the well known Q-learning algorithm. 
We start by defining a general algorithm of Q-learning with experience replay, without specifically defining the replay sampling scheme. This gives a general framework which can be used for various purposes. We then prove sufficient conditions on the replay scheme for convergence of the Q values. Finally, we will use this framework with a particular replay scheme to achieve a safe policy. We start by describing the algorithm.

Let M be an MDP = $<S,A,P,R,\gamma>$. 
For each transition $s,a,s',r$ there exists some probability, induced by $P, R$, to observe the transition from state $s$ to state $s'$ with reward $r$, given an action $a$. We call this probability $q(s',r|s,a)$.
We also assume that for each transition $(s,a,s',r)$, at each iteration $t > 1$, there exists a replay probability $w_t(s,a,s',r)$ which gives the probability to replay the transition. 
 
At each iteration $t$, with probability $v$ the Q values are updated with a sample from the replay buffer, where the sample is chosen using the replay probability function $w_t$. With probability $1-v$, the Q values are updated with a sample from the MDP (See algorithm~\ref{alg:Q-learning with replay}), and the transition is stored in the memory buffer. $w_t$, the replay probability function, is then updated using a function $f$ that depends on the current replay probability function $w_t$, and past experiences $F_t = (s_1, a_1, r_1, s_2,...,r_t)$.
Note that we allow a constant initial exploration time $T_0$ in order to ensure we have a valid replay probability function ($T_0 \geq 1$ since we start with an empty memory buffer and a zero probability for replay at the first iteration). 
\subsection{Convergence of Q-Learning with convergent replay}
We now turn to prove the convergence of the above algorithm. We show that if the replay probability function obeys to a very intuitive criteria, namely, that the sequence of replay probability functions $w_0,w_1,w_2,...,$ converges to some function $w^\infty$, then algorithm~\ref{alg:Q-learning with replay} converges. We sketch here the proof and give the full details in the appendix. 

We use the framework of stochastic iterative algorithms, and show how our algorithm can be written as a stochastic iterative algorithm with a time-varying contraction mapping which depends on the replay probability function at every iteration. 
We prove first that under the condition that the sequence of replay probability functions converges, the sequence of time-varying contraction mappings converges as well. Then, we prove the convergence of any stochastic iterative algorithm with a sequence of time-varying contraction mappings, even ones that have different fixed points, as long as the sequence of contraction mappings converges. Finally, we show that our algorithm is a stochastic iterative algorithm with a convergent sequence of contraction mappings, which gives the convergence result. 
\paragraph{Stochastic Iterative Algorithms}
A stochastic iterative algorithm is an iterative algorithm of the form:
$x_{t+1}(i) = (1- \alpha_t(i)) x_t(i) + \alpha_t(i) \cdot \left( (H_t x_t)(i) + e_t(i) \right) $ where $H_t$ are maximum norm contraction mappings with some contraction factor $\gamma < 1$, $\alpha_t$ is a step size, and $e_t$ is a noise term. For such algorithms, under the following assumptions, $x_t$ converges w.p. 1:
\begin{itemize}
\item For all t,i, the step sizes satisfy: $\alpha_t(i) \geq 0$,
$\sum_{t=0}^\infty\alpha_t(i) = \infty, \mbox{   } \sum_{t=0}^\infty \alpha_t^2(i) < \infty$
\item For every t,i the noise term $e_t(i)$ satisfy: \\
$\mathbb{E}[e_t(i)|F_t] = 0$, there exists a constant $K$ s.t. $\mathbb{E}[e_t^2(i)|F_t] \leq K (1 + \|x_t \|^2 )$
\end{itemize}
\begin{algorithm}[t]
\caption{Q-learning with convergent experience replay}
\label{alg:Q-learning with replay}
\begin{algorithmic}
\STATE input: $\gamma\in (0,1) =$ decay parameter, $\alpha_t(\cdot, \cdot)$ = learning rate function, $v$ = probability for replay, $f$ = weight update function, $T_0 \geq 1$ = initial exploration time
\STATE init: $\forall a \in A, s \in S: Q_0(s,a) = c_0, t=0, \forall s,s' \in S, a\in A, r \in [r_{min}, r_{max}]: w_t(s,a,s',r) = 0$
\FOR{$t = 1,2,..., T_0$}{ 
\STATE Explore and update the Q-values:
$Q_t(s_t,a_t) = (1-\alpha_t(s_t,a_t)) Q_{t-1}(s_t,a_t) + \alpha_t(s_t,a_t) \left( r_t + \gamma  max_{a'} Q_{t-1}(s_{t+1},a') \right)$
\STATE Store transition in memory: $s_t, a_t, r_t, s_{t+1}$
\STATE Update probability vector $w_{t+1} = f(w_t, F_t)$
}
\ENDFOR
\FOR{$t = T_0 + 1,T_0 + 2,...$}{
\STATE With probability $v$ perform a replay iteration:
\INDSTATE Choose a replay sample $(s,a,s',r)$ according to the probability vector $w_t$\\
\INDSTATE Update: 
\INDSTATE $Q_t(s,a) = (1-\alpha_t(s,a)) Q_{t-1}(s,a) + \alpha_t(s,a) \left( r + \gamma \cdot max_{a'} Q_{t-1}(s',a') \right)$
\STATE With probability $1-v$, perform an iteration in the MDP: \\
\INDSTATE Choose action $a_t = argmax_a (Q_{t-1}(s_t,a))$ (or explore)\\
\INDSTATE Receive $r_t$, transition to $s_{t+1}$
\INDSTATE Update: 
\INDSTATE $Q_t(s_t,a_t) = (1-\alpha_t(s_t,a_t)) Q_{t-1}(s_t,a_t) + \alpha_t(s_t,a_t) \left( r_t + \gamma  max_{a'} Q_{t-1}(s_{t+1},a') \right)$\\
\INDSTATE Store transition in memory: $s_t, a_t, r_t, s_{t+1}$ \\
\STATE Update probability vector $w_{t+1} = f(w_t, F_t)$
}
\ENDFOR
\end{algorithmic}
\end{algorithm}
In order to show that algorithm~\ref{alg:Q-learning with replay} is a stochastic iterative algorithm, we start by defining a suitable mapping for our algorithm, which we will later show is a maximum norm contraction mapping.
Since in every iteration, with probability $v$ we sample from the replay buffer using a replay probability function $w_t$, and  with probability $1-v$ we sample from the MDP, we need to account for these two types of samples in our mapping. Thus, we start by defining an alternative probability transition function $\tilde{p}_t$ and reward distribution $\tilde{R}_t$ which will account for both sample types. Notice that $\tilde{p}_t$ and $\tilde{R}_t$ are time dependent since at every iteration we have a different replay probability function $w_t$. 
\begin{align}
&\tilde{p}_t(s'|s,a) := \sum_{r = r_{min}}^{r_{max}} (1-v) \cdot \phi_{1,t}(s,a,s',r) + v \cdot \phi_{2,t}(s,a,s',r)  \label{p_t} \\
&\tilde{R}_t(s,a,s') := \sum_{r = r_{min}}^{r_{max}} r \frac{ (1-v) \phi_{1,t}(s,a,s',r) + v  \phi_{2,t}(s,a,s',r) }{\sum\limits_{\bar{r} = r_{min}}^{r_{max}}  (1-v)  \phi_{1,t}(s,a,s',\bar{r}) + v  \phi_{2,t}(s,a,s',\bar{r})} \label{R_t}
\end{align}
where:
\begin{align}
&\phi_{1,t}(s,a,s',r) := \frac{q(s',r|s,a)}{\sum\limits_{\bar{s} \in S} \sum\limits_{\bar{r} = r_{min}}^{r_{max}} (1-v) \cdot q(\bar{s},\bar{r}|s,a) + v \cdot w_t(s,a,\bar{s},\bar{r})} \label{phi1} \\
&\phi_{2,t}(s,a,s',r) := \frac{w_t(s,a,s',r)}{\sum\limits_{\bar{s} \in S} \sum\limits_{\bar{r} = r_{min}}^{r_{max}} (1-v) \cdot q(\bar{s},\bar{r}|s,a) + v \cdot w_t(s,a,\bar{s},\bar{r})} \label{phi2} 
\end{align}
Notice that if $w_t(s,a,\cdot,\cdot) = 0$, i.e. there is no probability to perform replay on a state-action pair $(s,a)$, then we result back to the original transition probabilities and expected reward as in the MDP, i.e. $\tilde{p}_t(s,s'|a) = p(s,s'|a), \tilde{R}_t(s,a,s') = R(s,a,s')$. \\ 
We now define the mapping:
\begin{align*}
&H_tQ(s,a) = \sum_{s' \in S} \tilde{p}_t(s'|s,a) \cdot \left( \tilde{R}_t(s,a,s') + \gamma \cdot max_{b \in A} Q(s',b) \right)
\end{align*}

The following lemma states that if the sequence of replay probability functions  $w_0,w_1,w_2,...$ converges uniformly to some function $w^\infty$, then the sequence of mappings $H_0, H_1, H_2,...$ also converges uniformly to $H^\infty$. Furthermore, if $H_0, H_1, H_2,...$ are all contraction mappings, than the sequence of fixed points also converges.
\begin{lemma}\label{lemma:convergence_of_mappings}
Let $w_i: S \times A \times S \times \mathbb{R} \rightarrow \mathbb{R}$ be probability functions for $i=0,1,2...$, and let $w^\infty: S \times A \times S \times \mathbb{R} \rightarrow \mathbb{R} $. 
Define:
\begin{align*}
&H_tQ(s,a) = \sum_{s' \in S} \tilde{p_t}(s'|s,a) \cdot \left( \tilde{R_t}(s,a,s') + \gamma \cdot max_{b \in A} Q(s',b) \right)
\end{align*}
with $\tilde{p_t}, \tilde{R_t}$ as defined in equations~\ref{p_t},~\ref{R_t},
and similarly for $H^\infty Q(s,a), \tilde{p}^\infty(s'|s,a), \tilde{R}^\infty(s,a,s')$ with $w^\infty$. \\ 
If the sequence $\{w_t\}_{t=0}^\infty$ converges uniformly to $w^\infty$, then $\{H_t\}_{t=0}^\infty$ converge uniformly to $H^\infty$.
Furthermore, if $\{H_t\}_{t=0}^\infty$ and $H^\infty$ are contraction mappings with fixed points $\{Q^\ast_t\}_{t=0}^\infty$ and $Q^{\ast,\infty}$ correspondingly, then the sequence $\{Q^\ast_t\}_{t=0}^\infty$ converges to $Q^{\ast,\infty}$.
\end{lemma}
The proof of the above lemma is a simple application of the limit rules of convergent sequences. The full proof is found in appendix~\ref{app:convergence_of_mappings_proof}.
Next, we show that the mappings defined above are all contraction mappings.
\begin{lemma}\label{lemma:contraction_and_fixed_point_of_mappings}
Define $H_tQ(s,a)$ as:
\begin{align*}
&H_tQ(s,a) = \sum_{s' \in S} \tilde{p_t}(s'|s,a) \cdot \left( \tilde{R_t}(s,a,s') + \gamma \cdot max_{b \in A} Q(s',b) \right)
\end{align*}
with $\tilde{p_t}, \tilde{R_t}$ as defined in equations~\ref{p_t},~\ref{R_t}.\\
Let $\gamma \in (0,1)$, $v \in [0,1)$, and $\sum_{s \in S} \sum_{s' \in S}\sum_{a \in A} \sum_{r=r_{min}}^{r_{max}} w_t(s,s',a,r) = 1$.
Then, for all $t$, $H_t Q(s,a)$ is a maximum norm contraction mapping with contraction $\gamma$.
\end{lemma}
The full proof is found in appendix~\ref{app:contraction_and_fixed_point_of_mappings_proof}, and is based on the fact that $\sum_{s' \in S} \tilde{p_t}(s'|s,a) = 1 $. This, together with the definition of $H_t$, give the result. 

We now prove that a stochastic iterative algorithm with a \emph{convergent} sequence of contraction mappings converges, extending the proof for stochastic iterative algorithms with time-dependent mapping that share a fixed point (\cite{10.5555/560669}):
\begin{lemma} \label{lemma:stochastic_iterative_algorithm_with_convergent_mapping}
Let $x_t$ be the sequence generated by the following iteration:
\begin{align*}
x_{t+1}(i) = (1- \alpha_t(i)) x_t(i) + \alpha_t(i) \cdot \left( (H_t x_t)(i) + e_t(i) \right) 
\end{align*}
where $\{H_t\}_{t=0}^\infty$ is a convergent sequence of maximum norm contraction mappings with contraction factor $\gamma < 1$, s.t. $lim_{t \rightarrow \infty} H_t = H^\ast$, and for each $t$, $H_t$ has a fixed point $x^\ast_t$. Assume $H_t$ is chosen without knowledge of the future (i.e. it is a function of the history at iteration t, $F_t$). Let $x^\ast$ be the fixed point of $H^\ast$.\\
We assume the following:
\begin{enumerate}
\item For all t,i, the step sizes satisfy: $\alpha_t(i) \geq 0$, $\sum_{t=0}^\infty\alpha_t(i) = \infty$, $\sum_{t=0}^\infty \alpha_t^2(i) < \infty$
\item For every t,i the noise term $e_t(i)$ satisfy: 
\begin{align*}
&\mathbb{E}[e_t(i)|F_t] = 0, \mbox{ there exists a constant } K \mbox{ s.t. } \mathbb{E}[e_t^2(i)|F_t] \leq K (1 + \|x_t \|^2 )
\end{align*}
\end{enumerate}
Then, $x_t$ converges to $x^\ast$ with probability 1.
\end{lemma}
The proof of the above lemma is found in appendix~\ref{app:stochastic_iterative_algorithm_with_convergent_mapping_proof}. We use proposition 4.7 in~\cite{10.5555/560669} that gives a convergence result for stochastic iterative algorithms with additional noise terms. We define an additional noise term $u_t(i) = (H_t x_t)(i) - (H^\ast x_t)(i)$, and show that it has the property that there exists a non-negative random sequence $\theta_t$ that converges to zero with probability 1, and is such that:
$|u_t(i)| \leq \theta_t \cdot (\|x_t\| + 1), \forall i,t$. This gives us the result.

Finally, we show that algorithm~\ref{alg:Q-learning with replay}, when using a convergent sequence of replay probability functions, can be written as an stochastic iterative algorithm with the convergent sequence of contraction mappings as defined above. In algorithm~\ref{alg:Q-learning with replay}, we use the function $f$ to update the replay probability vector $w_t$ as a function of the history $F_t = \{s_0,a_0,r_0,s_1,a_1,r_1,...s_t,a_t,r_t\}$. We now assume that the resulting sequence of functions $\{ w_t \}_0^\infty$ converges to a function $w^\infty$.

\begin{theorem} \label{lemma:Q_learning_with_as_iterative_stochastic_algorithm}
Let $f$ be a function of $w_t$ and $F_t = \{s_0,a_0,r_0,s_1,a_1,r_1,...s_t,a_t,r_t\}$, and assume that the resulting sequence of functions $\{ w_t \}_0^\infty$ converges to a function $w^\infty$, then, algorithm~\ref{alg:Q-learning with replay} converges with probability 1 given the following:
\begin{itemize}
\item For all t,i, the step sizes satisfy: $\alpha_t(i) \geq 0$, 
$\sum_{t=0}^\infty\alpha_t(i) = \infty$,$\sum_{t=0}^\infty \alpha_t^2(i) < \infty$
\item GLIE (greedy in the limit with infinite exploration)- For every state $s \in S$ that is visited infinitely often, each action in that state is chosen infinitely often with probability 1.  
\end{itemize}
\end{theorem}
The proof follows by re-writing the Q-values update using the contraction mappings $H_t$ as defined above, and defining the error term: 
\begin{align*}
&e_t(s_t,a_t) = r_t(s_t,a_t,s_{t+1}) + \gamma \cdot max_{b \in A} Q_t(s_{t+1},b) \\
&- \sum_{s' \in S} \tilde{p}_t(s'|s_t,a_t) \left( \tilde{R}_t(s_t,a_t,s') + \gamma \cdot max_{b \in A} Q_t(s',b) \right)
\end{align*}
We show that $E[e_t(s_t,a_t)|F_t] = 0$ and $E[e_t^2(s_t,a_t)|F_t] < D$ for some constant $D$, which gives us the required conditions for convergence in lemma~\ref{lemma:stochastic_iterative_algorithm_with_convergent_mapping}.
The full proof is found in appendix~\ref{app:Q_learning_with_as_iterative_stochastic_algorithm_proof}. 


\section{Replay for safety}
We now show how our framework can be used for safe RL. Consider the task of learning a safe policy in terms of the variance in the immediate reward. We show that such a policy can be learned using a suitably designed replay mechanism. 
We design a replay scheme in which transitions of state-action pairs with higher variance have a higher probability for replay, and in which lower rewards in those state-action pairs have a higher probability for replay. This ensures that state-action pairs with higher variance have more replay iterations, and in these iterations, worse outcomes are replayed, creating a negative bias that leads the policy to choose different actions and avoid high variance state-action pairs that have very bad potential outcomes. We start by giving an example of a replay scheme for safe RL, and then generalize this to a family of replay schemes, all leading to a safe policy.
\subsection{Variance prioritized replay}
We propose a replay scheme that takes into account the empirical variance of the immediate reward of each state-action pair, and selects experiences from the replay buffer relatively to the state-action pairs' reward variance, and with a bias to negative rewards. Denote by $var_t(r(s,a))$ the empirical variance at time $t$ of the reward received after taking action $a$ in state $s$. 
We propose the following replay scheme, and prove that it converges.
\begin{figure}
\centering
\includegraphics[scale=0.37]{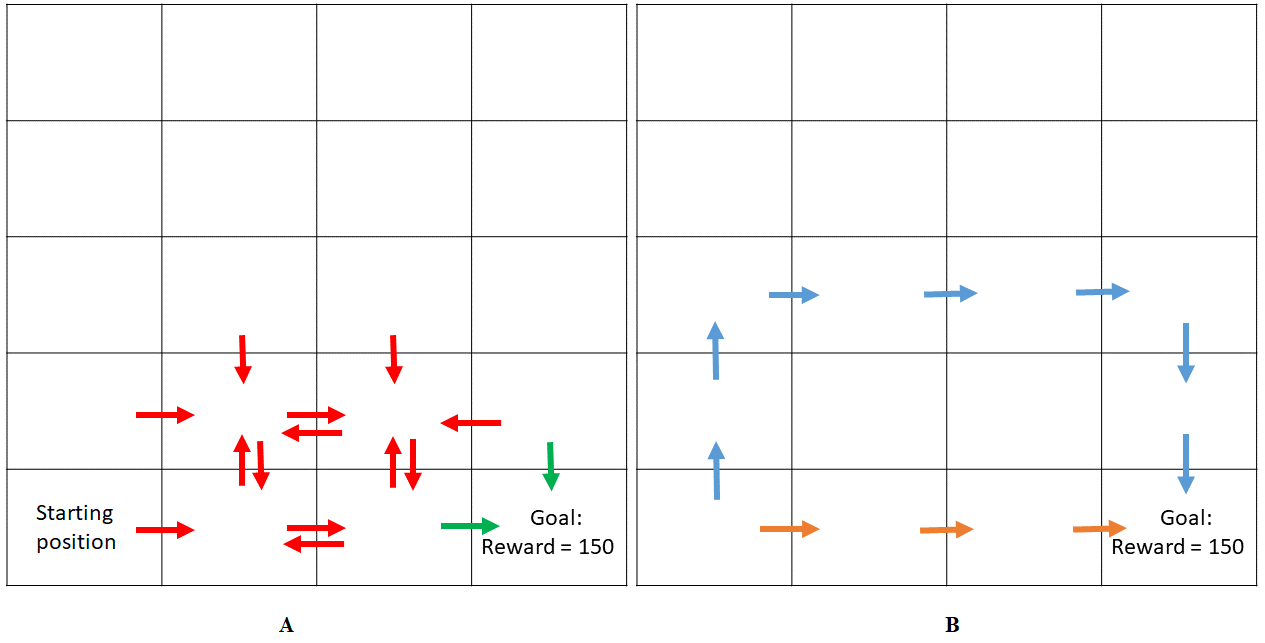}
\caption{A. Grid of first experiment. All transitions marked in red are volatile transitions in which the reward is high ($100$) with probability $0.6$ and low ($-100$) with probability $0.4$. Transitions to the goal position which are marked in green result in a reward of $150$. B. Orange: Policy of the Q-learning algorithm. Blue: Policy of algorithm~\ref{alg:Q-learning with replay} with replay scheme from lemma~\ref{lemma:variance_based_replay_convergence}. the optimal action of the policy is drawn by an arrow.}
\label{fig:exp1_environment}
\end{figure}
\begin{lemma} \label{lemma:variance_based_replay_convergence}
Let $w_0(s,a,s',r) = 0$ for all $s,a,s',r$. Denote by $M_t(r(s,a))$ the set of unique rewards seen in transitions from state $s$ with action $a$ until iteration $t$, and $L_t(s,a,r)$ the set of unique next-states seen in transitions from state $s$ with action $a$ and reward $r$ until iteration $t$.
Denote by $S_r(s,a) := \{s' \in S : Pr(r(s,a,s') = r) > 0 \}$. \\
For each $t$, let:
\begin{align*}
& w_t(s,a,s',r) = \frac{var_t(r(s,a))}{\sum\limits_{\bar{s} \in S} \sum\limits_{\bar{a} \in A} var_t(r(\bar{s},\bar{a})))} \cdot \frac{e^{-\beta  r}}{\sum\limits_{r' \in M_t(r(s,a))} e^{-\beta  r'}} \cdot \frac{1}{|L_t(s,a,r)|}
\end{align*}
if $s' \in L_t(s,a,r)$ and $r \in M_t(r(s,a))$, and $w_t(s,a,s',r)=0$ otherwise.
Then, under the GLIE assumption, $w_t$ converges to $w^\infty$ a.s., i.e.
$\lim_{t \rightarrow \infty} w_t = w^\infty$
where 
\begin{align*}
&w^\infty(s,a,s',r) = 
\begin{cases}
 \frac{var(r(s,a))}{\sum\limits_{\bar{s} \in S} \sum\limits_{\bar{a} \in A} var(r(\bar{s},\bar{a})))}  \frac{e^{-\beta  r}}{\sum\limits_{r' = r_{min}}^{r_{max}} e^{-\beta  r'}}  \frac{1}{|S_r(s,a)|} & \text{if $s' \in S_r(s,a)$} \\
0 & \text{o.w.}
\end{cases} 
\end{align*} 
\end{lemma}
The full proof is found in appendix~\ref{app:variance_based_replay_convergence_proof}, and is based on the fact that as we sample more and more transitions from the MDP, the sample variance converges to the true variance of the reward distribution, and the minimal and maximal reward that was observed converges to the minimal and maximal reward of the distribution. Thus, in the limit of $t \rightarrow \infty$ and using the GLIE assumption (which ensures infinite number of visits in all state-action pairs), we get that the sequence of replay probability functions converges. 
\subsection{Safe experience replay}
We formulate the intuition of replaying highly variable state-action pairs in a biased manner in assumption~\ref{assumption:replay_variance}. We show that replay schemes that satisfy this assumption allow to learn a safe policy. 
Note that $w^\infty$ in the replay scheme proposed in lemma~\ref{lemma:variance_based_replay_convergence} is an example of a replay scheme that satisfies assumption~\ref{assumption:replay_variance}, and thus theorem~\ref{lemma:safe_policy_after_replay} holds for it. 
\begin{assumption} \label{assumption:replay_variance}
Let $w$ be a replay probability function (depending only on the MDP parameters) which satisfies the following:
\begin{itemize}
\item state-action pairs with higher variance have a higher replay probability, i.e, if $var(r(s,a)) > var(r(s',a'))$ then $w(s,a,\cdot,r) > w(s',a',\cdot,r)$. 
\item For each state-action $(s,a)$, transitions to states $s'$ with lower reward value have a higher replay probability, i.e. if $r < r'$ then $w(s,a,\cdot,r) > w(s,a,\cdot,r')$. 
\item For all state-action pairs $(s,a)$ and $s \in S_{r_{min}}(s,a)$, the following holds: 
\begin{align*}
&\Lim{\frac{var(r(s,a))}{\sum_{s \in S, a \in A}var(r(s,a))} \rightarrow 1} w(s,a,s',r_{min}) = \frac{1}{|S_{r_{min}}(s,a)|}
\end{align*}
where $S_{r_{min}}(s,a) := \{s' \in S : Pr(r(s,a,s') = r_{min}) > 0 \}$.
\end{itemize}
\end{assumption}
The third bullet in assumption~\ref{assumption:replay_variance} intuitively means that $w$ concentrates most of the probability on highly variable state-action pairs, such that only those pairs are effected by the biased replay.\\  
We inspect the behavior of $Q^\ast(s,a)$ in the limit of $\frac{var(r(s,a))}{\sum_{s \in S, a \in A}var(r(s,a))} \rightarrow 1$, i.e., for state-action pairs with especially high variance. We show that the resulting Q value will be lower than the Q values of such state-action pairs in a setting where $v=0$, i.e. when no replay is done, and that this can lead to a safer policy that chooses a less variable action.
\begin{theorem} \label{lemma:safe_policy_after_replay}
Denote by $Q_{v=0}^\ast$ the optimal Q-values in the MDP when no replay is performed, i.e. when $v=0$. 
For any choice of the MDP parameters, if $\frac{var(r(s_i,a_i))}{\sum_{s \in S, a \in A}var(r(s,a))}$ is sufficiently close to 1 and $w^\infty$ is a probability replay function satisfying assumption~\ref{assumption:replay_variance}, then for any choice of $\{w_t\}_{t=1}^\infty$ converging to $w^\infty$, for $(s_i,a_i)$ such that $\forall b \neq a_i \in A: Q_{v=0}^\ast(s_i,b) < Q_{v=0}^\ast(s_i,a_i)$, if there exists an action $b \in A$ s.t.:
\begin{align*}
& Q_{v=0}^\ast(s_i,a_i) - Q_{v=0}^\ast(s_i,b) < v \cdot \left(  R(s_i,a_i) - r_{min} \right) \\
& + \gamma \sum_{s' \in S} p(s'|s_i,a_i) \left( \max_{b \in A} Q_{v=0}^\ast(s',b) - \max_{b \in A} Q^\ast(s',b)\right) \\
& + v \gamma \cdot \left( \sum_{s' \in S}  p(s'|s_i,a_i) \cdot \max_{b \in A} Q^\ast(s',b) - \sum_{s' \in S_{r_{min}}} \frac{1}{|S_{r_{min}}(s_i,a_i)|} \cdot \max_{b \in A} Q^\ast(s',b) \right)
\end{align*}
Then,
$Q^\ast(s_i,a_i) < Q^\ast(s_i,b)$
\end{theorem}
The above theorem shows that an action that was optimal in the regular setting, when only expected return is maximized and no replay is performed, e.g., $a_i$, is no longer the action chosen when performing replay for safety, and the optimal action becomes $b$, an action with lower variance.
Intuitively, the bound on $Q_{v=0}^\ast(s_i,a_i) - Q_{v=0}^\ast(s_i,b)$ in the theorem tells us that for the policy to switch the optimal action, there has to be an action $b$ for which the distribution of rewards and next-states is higher than the gap between $Q_{v=0}^\ast(s_i,a_i)$ and $Q^\ast(s_i,a_i)$, created by the replay. If no such action $b$ exists (and $\pi(s_i)$ remains $a_i$), that means that the variance in $(s_i,a_i)$ was not due to very negative outcomes but due to variance of relatively positive outcomes (relative to the other actions), and thus not switching actions won't lead to dangerous situations.
The full proof of theorem~\ref{lemma:safe_policy_after_replay} is found in appendix~\ref{app:safe_policy_after_replay_proof}. The basic idea is that when extensive replay is performed for some state-action pair which incurs high variance, then the effective distributions of immediate reward and next-state are biased in such a way that the learned Q-values are lower. This is because we give high replay probability for the most negative outcomes of such state-action pairs, thus creating a seemingly high probability for a very bad outcome. This leads to a change in the policy if the Q-values are shifted sufficiently.

\section{Experiments}
\label{sec:experimens}
\begin{figure}
\centering
\includegraphics[scale=0.43]{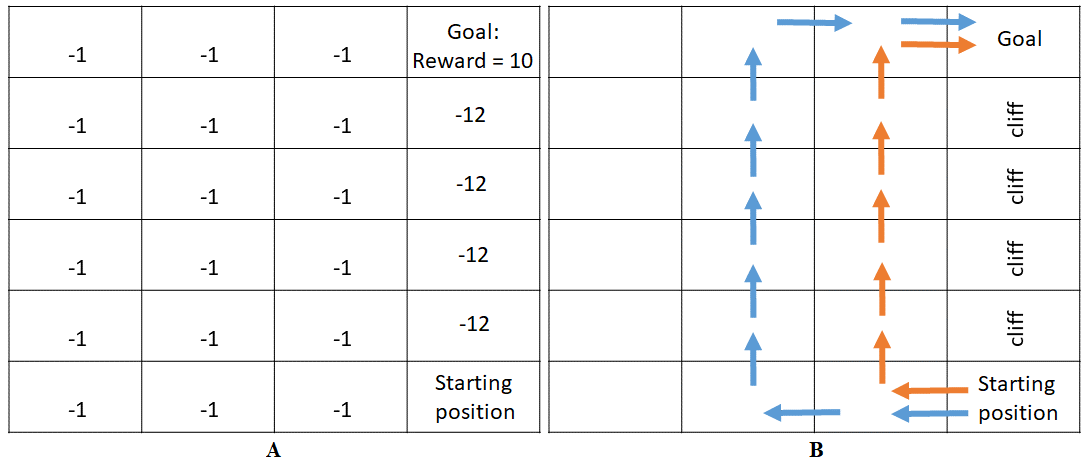}
\caption{A. Grid of second experiment- the cliff environment. All transitions incur a cost of $-1$ except transitions to the goal position (reward of $10$) and transitions that fall off the cliff (cost of $-12$). All Actions succeed w.p. $0.9$ and fail w.p. $0.1$ (in which case a transition to a random direction is done), except transitions that lead to the goal position. B. Orange: Policy of the Q-learning algorithm. Blue: Policy of algorithm~\ref{alg:Q-learning with replay} with replay scheme from lemma~\ref{lemma:variance_based_replay_convergence}. The optimal action of the policy is drawn by an arrow.}
\label{fig:exp2_experiement}
\end{figure}
We demonstrate how the replay scheme from lemma~\ref{lemma:variance_based_replay_convergence}, used with algorithm~\ref{alg:Q-learning with replay}, can result in a safe policy that avoids catastrophic events in the cost of a slightly lower expected return.
This in fact provides experimental support to the limiting behavior we investigate in theorem~\ref{lemma:safe_policy_after_replay}.
We test algorithm~\ref{alg:Q-learning with replay} in two grid world environments (code is provided in supplementary). In the first, some transitions result in an immediate reward that has high variance. In this environment, the next-state transitions are deterministic, i.e. the variance stems from variance in the reward distribution. In the second environment, the variance stems from high variance in the next-state probability distribution, whereas the immediate reward is deterministic. 
In both grid environments, each grid cell is a state and the possible actions in each state are moving in one of the 4 directions- up, down, left or right. In case the action chosen results in stepping outside of the grid size (for instance, a left in the left border of the grid), then no movement is done. 
We use the following parameters in both environments - $v=0.5, \beta = 5$. 
For learning, we run for $50000$ episodes, where in each episode the starting position is chosen at random. We use the learning rate $\frac{1}{t}^{0.6}$. 
Figure~\ref{fig:exp1_environment}a graphically describes the reward distribution in the first environment. Here, we use a 4 by 5 grid. The grid contains state-action pairs for which the immediate reward is $-100$ with probability $0.4$ and $100$ with probability $0.6$ (transitions marked with a red arrow). For all other state-action pairs, the reward is $-1$, which creates the motivation to use shorter paths to reach the goal. The reward for reaching the goal position (i.e. the reward received for all transitions that lead to the goal position) is $150$ (transitions marked with a green arrow). The episode ends when the goal is reached.  
In this environment, the optimal path in terms of the expected return is taking the shortest path to the goal, i.e. taking the 'right' action until reaching the goal position. A safe policy, however, should overturn the highly variable states and take a slightly longer but less dangerous path to the goal. 
Figure~\ref{fig:exp1_environment}b show the resulting policy of Q-learning (in orange) and of our algorithm (in blue). Indeed, the resulting policy in case replay is performed is a safe policy, whereas the regular Q-learning agent learns to walk the more dangerous (and with higher expected return) path.
The second environment is a variation on the cliff environment from \cite{sutton2018reinforcement}. Here, variance stems from non-deterministic transitions. We use a grid of size 4 by 6 where every action succeeds with probability $0.9$ and fails with probability $0.1$. If an action fails, a random direction is chosen and the transition is performed to that direction. All state-action pairs except those leading to the goal position have a failure probability, and transitions to the goal position succeed with probability $1$. The cliff environment contains a cliff - falling off the cliff results in a high negative reward of $-12$. Reaching the goal results in a high positive reward of $10$. Any other transition results in a reward of $-1$. An optimal agent (in terms of expected reward) would walk as closely as possible to the edge of the cliff, so that the shortest path to the goal is taken. A safe agent, however, would take a longer but safer path and walk farther from the cliff edge. Figure~\ref{fig:exp2_experiement} portrays the environment and the resulting policies. As expected, Q-learning leads to the optimal and most dangerous policy, and our algorithm leads to a safer policy. 
We also run both the algorithms proposed in~\cite{mihatsch2002risk} and~\citep{gaskett2003reinforcement} on our two environments. ~\citet{mihatsch2002risk} propose a Q-learning algorithm in which the TD errors are multiplied in an asymmetric manner in order to emphasize negative TD errors over positive ones. Positive TD errors are multiplied by $1-\kappa$, and negative ones by $1+\kappa$, where $\kappa \in (-1,1)$ determines the extent to which the policy is risk averse (or risk-seeking). The algorithm successfully learned a safe policy in the first environment, but failed to do so in the second environment (the cliff environment), for any $\kappa$ value between 0 and 1 with a step size of 0.01, with $500000$ episodes. All policies attained were either overly pessimistic and jumped off the cliff immediately (a well documented phenomena in this environment, see~\citep{gaskett2003reinforcement}), or learned to walk the dangerous path alongside the cliff, as does the regular Q-learning. We suspect that this is due to the distribution of TD-errors, which is a multimodal distribution with distinct peaks in $0, -100, 100$ in the first environment, and a unimodal, wider distribution in the second environment, which is more difficult to manipulate in order to obtain an exact balance between pessimism (risk-aversion) and optimism (optimizing the expected return). \\
\citet{gaskett2003reinforcement} use the worst-case criteria, as outlined in section~\ref{sec:prev_safe_RL}, to obtain a policy that achieves a maximum expected return under the worst trajectory possible for that policy. Here, a safe policy is attained in the second environment, but not in the first. In fact, any algorithm optimizing this criteria cannot attain a safe policy in the first environment since all transitions are deterministic and thus all trajectories are exactly those chosen by the policy (there is no probability to reach a state not intended by the policy, by randomness of the environment), and the variance stems only from the reward distribution. Thus, under this criteria, the resulting policy will be exactly the one obtained by regular Q-learning. 
In conclusion, we show that our algorithm is able to learn a safe policy in environments with different sources of variability, even in situations where other methods fail.
\section{Conclusions}
In this work we showed how designing a replay probability function in a suitable way may lead to desirable properties of the resulting policy, such as safety. We give a clear and intuitive condition on the sequence of replay probability functions, namely, that it converges, that ensures convergence of the algorithm. Such a mechanism allows us to easily manipulate the effective next-state and reward distributions as seen by the algorithm, in such a way that no modification of the optimality criteria is needed in order to get to a policy with different traits. This is a powerful tool on the one hand, but it is simple and can easily be incorporated into many RL algorithms on the other hand. Since experience replay is very commonly used in RL, this is a natural new way of using this heuristics.
Here, we define a specific type of replay probability scheme to lead the algorithm to learn a safe policy. We prove that the the limiting behavior of the policy will be to choose less variable state-action pairs, and finally, we support our theoretical findings with experimental evidence. We believe that the idea of using specifically crafted replay schemes as a measure of shaping the resulting policy is very powerful. It can be used in the same manner for different situations, all without changing the algorithm itself.  
\newpage
\bibliography{bib_replay_for_safety}
\bibliographystyle{plainnat}

\newpage
\onecolumn
\section{Appendix}
\label{sec:proofs}
\subsection{Proof of Lemma~\ref{lemma:convergence_of_mappings}}
\label{app:convergence_of_mappings_proof}
\begin{proof}
Since $\{w_t\}_{t=0}^\infty$ converges uniformly to $w^\infty$, $\{\phi_{1,t}\}_{t=0}^\infty$ and $\{\phi_{2,t}\}_{t=0}^\infty$ also converge to $\phi_1^\infty := \frac{q(s',r|s,a)}{\sum_{\bar{s} \in S} \sum_{\bar{r} = r_{min}}^{r_{max}} (1-v) \cdot q(\bar{s},\bar{r}|s,a) + v \cdot w^\infty(s,a,\bar{s},\bar{r})}$ and $ \phi_2^\infty := \frac{w^\infty(s,a,s',r)}{\sum_{\bar{s} \in S} \sum_{\bar{r} = r_{min}}^{r_{max}} (1-v) \cdot q(\bar{s},\bar{r}|s,a) + v \cdot w^\infty(s,a,\bar{s},\bar{r})}$ correspondingly, using the limit laws of convergent sequences. For the same reason, $R\tilde{R}_t, \tilde{p}_t$ also converge to $\tilde{R}^\infty$ and $\tilde{p}^\infty$, which are defines as before, but with $\phi_1^\infty, \phi_2^\infty$. We use the limit laws of convergent sequences again to obtain:
\begin{align*}
&\lim_{t \rightarrow \infty} H_t Q(s,a)  \\
& = \lim_{t \rightarrow \infty} \sum_{s' \in S} \tilde{p_t}(s'|s,a) \cdot \left( \tilde{R_t}(s,a,s') + \gamma \cdot max_{b \in A} Q(s',b) \right)  \\
& = \sum_{s' \in S} \lim_{t \rightarrow \infty}  \tilde{p_t}(s'|s,a) \cdot \left( \lim_{t \rightarrow \infty} \tilde{R_t}(s,a,s') + \gamma \cdot max_{b \in A} Q(s',b) \right) \\ 
& = \sum_{s' \in S} \tilde{p}^\infty(s'|s,a) \cdot \left( \tilde{R}^\infty(s,a,s') + \gamma \cdot max_{b \in A} Q(s',b) \right) \\
& = H^\infty Q(s,a)
\end{align*} 

If $\{H_t\}_{t=0}^\infty$ are all contraction mappings with fixed points $\{Q^\ast_t\}_{t=0}^\infty$, we use theorem 2 in \cite{nadler1968sequences} to get that the sequence of fixed points $\{Q^\ast_t\}_{t=0}^\infty$ converges to the fixed point of $H^\infty$, i.e. $Q^{\ast,\infty}$.

\end{proof}

\subsection{Proof of Lemma~\ref{lemma:contraction_and_fixed_point_of_mappings}}
\label{app:contraction_and_fixed_point_of_mappings_proof}
\begin{proof}
For any two vectors $Q, \bar{Q}$, we have:
\begin{align*}
& \left| H_t Q(s,a) - H_t \bar{Q}(s,a)\right| \\
& = \left| \sum_{s' \in S} \tilde{p_t}(s'|s,a) \cdot \left( \tilde{R_t}(s,a,s') + \gamma  max_{b \in A} Q(s',b) \right) - \left( \sum_{s' \in S} \tilde{p_t}(s'|s,a) \cdot \left( \tilde{R_t}(s,a,s') + \gamma  max_{b \in A} \bar{Q}(s',b) \right) \right) \right| \\
& \leq \sum_{s' \in S} \tilde{p_t}(s'|s,a) \cdot  \left| \left( \tilde{R_t}(s,a,s') + \gamma  max_{b \in A} Q(s',b) \right) - \left( \tilde{R_t}(s,a,s') + \gamma  max_{b \in A} \bar{Q}(s',b) \right) \right| \\
& \leq \gamma \cdot \sum_{s' \in S} \tilde{p_t}(s'|s,a) \cdot \left|  max_{b \in A} Q(s',b) - max_{b \in A} \bar{Q}(s',b) \right| \\
& \leq \gamma \cdot \sum_{s' \in S} \tilde{p_t}(s'|s,a) \cdot max_{b \in A} \cdot \left| Q(s',b) - \bar{Q}(s',b) \right| \\
& \leq \gamma \cdot \sum_{s' \in S} \tilde{p_t}(s'|s,a) \cdot max_{s \in S, b \in A} \cdot \left| Q(s,b) - \bar{Q}(s,b) \right|
\end{align*}
We now notice that $\sum_{s' \in S} \tilde{p_t}(s'|s,a) = 1 $, since: 
\begin{align*}
& \sum_{s' \in S} \tilde{p_t}(s'|s,a) = \sum_{s' \in S} \sum_{r = r_{min}}^{r_{max}} (1-v) \cdot \phi_{1,t}(s,a,s',r) + v \cdot \phi_{2,t}(s,a,s',r) \\
& = \sum_{s' \in S} \sum_{r = r_{min}}^{r_{max}} \frac{(1-v) \cdot q(s',r|s,a) + v \cdot w_t(s,a, s',r)}{\sum_{\bar{s} \in S} \sum_{\bar{r} = r_{min}}^{r_{max}} (1-v) \cdot q(\bar{s},\bar{r}|s,a) + v \cdot w_t(s,a,\bar{s},\bar{r})} \\
& = 1
\end{align*}
Thus we conclude:
\begin{align*}
& \left| H_t Q(s,a) - H_t \bar{Q}(s,a)\right| \\
& \leq \gamma \cdot max_{s \in S, b \in A} \cdot \left| Q(s,b) - \bar{Q}(s,b) \right| \\
& \leq \gamma \cdot \| Q -\bar{Q} \|_\infty
\end{align*}
Thus $H_t$ is a maximum norm contraction mapping with contraction factor $\gamma$.
\end{proof}

\subsection{Proof of Lemma~\ref{lemma:stochastic_iterative_algorithm_with_convergent_mapping}}
\label{app:stochastic_iterative_algorithm_with_convergent_mapping_proof}
\begin{proof}
We show how the sequence generated by the iteration above can be written as follows:
\begin{align*}
x_{t+1}(i) = (1- \alpha_t(i)) x_t(i) + \alpha_t(i) \cdot \left( (H^\ast x_t)(i) + e_t(i) + u_t(i) \right) 
\end{align*}
where $u_t(i)$ has the following property: there exists a non-negative random sequence $\theta_t$ that converges to zero with probability 1, and is such that:
$|u_t(i)| \leq \theta_t \cdot (\|x_t\| + 1), \forall i,t$ \\
This allows us to use proposition 4.7 in \cite{10.5555/560669}, and thus $x_t$ converges to $x^\ast$. \\
We define: $u_t(i) = (H_t x_t)(i) - (H^\ast x_t)(i)$. This gives us the iteration above, since we have:
\begin{align*}
x_{t+1}(i) &= (1- \alpha_t(i)) x_t(i) + \alpha_t(i) \cdot \left( (H^\ast x_t)(i) + e_t(i) + u_t(i) \right) \\
& = (1- \alpha_t(i)) x_t(i) + \alpha_t(i) \cdot \left( (H^\ast x_t)(i) + e_t(i) + (H_t x_t)(i) - (H^\ast x_t)(i) \right) \\
& = (1- \alpha_t(i)) x_t(i) + \alpha_t(i) \cdot \left((H_t x_t)(i) + e_t(i)\right)
\end{align*}
Now, since $\{H_t\}_0^\infty$ converges to $H^\ast$, we have that $ \forall x: lim_{t \rightarrow \infty}\ |H_t x - H^\ast x| = 0$, and thus there exists a non-negative sequence $\theta_t$ that converges to zero with probability 1, s.t. $|u_t(i)| \leq \theta_t \leq \theta_t \cdot (\|x_t\| + 1)$, and the proof is concluded. 
\end{proof}

\subsection{Proof of Theorem~\ref{lemma:Q_learning_with_as_iterative_stochastic_algorithm}}
\label{app:Q_learning_with_as_iterative_stochastic_algorithm_proof}
\begin{proof}
We show that Q-learning can be written as iterative stochastic algorithm with the properties specified in lemma~\ref{lemma:stochastic_iterative_algorithm_with_convergent_mapping}.
Define as before: 
\begin{align*}
HQ(s,a) = \sum_{s' \in S} \tilde{p}_t(s'|s,a) \cdot \left( \tilde{R}_t(s,a,s') + \gamma \cdot max_{b \in A} Q(s',b) \right)
\end{align*}
with:
\begin{align*}
& \phi_{1,t}(s,a,s',r) = \frac{q(s',r|s,a)}{\sum_{\bar{s} \in S} \sum_{\bar{r} = r_{min}}^{r_{max}} (1-v) \cdot q(\bar{s},\bar{r}|s,a) + v \cdot w_t(s,a,\bar{s},\bar{r})} \\
& \phi_{2,t}(s,a,s',r) = \frac{w_t(s,a,s',r)}{\sum_{\bar{s} \in S} \sum_{\bar{r} = r_{min}}^{r_{max}} (1-v) \cdot q(\bar{s},\bar{r}|s,a) + v \cdot w_t(s,a,\bar{s},\bar{r})} \\
&\tilde{p}_t(s'|s,a) = \sum_{r = r_{min}}^{r_{max}} (1-v) \cdot \phi_{1,t}(s,a,s',r) + v \cdot \phi_{2,t}(s,a,s',r) \\
&\tilde{R}_t(s,a,s') = \sum_{r = r_{min}}^{r_{max}} r \cdot \frac{ (1-v) \cdot \phi_{1,t}(s,a,s',r) + v \cdot \phi_{2,t}(s,a,s',r) }{\sum_{\bar{r} = r_{min}}^{r_{max}}  (1-v) \cdot \phi_{1,t}(s,a,s',\bar{r}) + v \cdot \phi_{2,t}(s,a,s',\bar{r})} 
\end{align*}
Define as well:
\begin{align*}
e_t(s_t,a_t) = r_t(s_t,a_t,s_{t+1}) + \gamma \cdot max_{b \in A} Q_t(s_{t+1},b) - \sum_{s' \in S} \tilde{p}_t(s'|s_t,a_t) \cdot \left( \tilde{R}_t(s_t,a_t,s') + \gamma \cdot max_{b \in A} Q_t(s',b) \right)
\end{align*}
We now show that $E[e_t(s_t,a_t)|F_t] = 0$ and $E[e_t(s_t,a_t)|F_t] < D$ for some constant $D$. \\
We look at: $E[r_t(s_t,a_t, s_{t+1}) + \gamma \cdot max_{b \in A} | F_t]$.
Define the random variable: 
\begin{align*}
X_t = \begin{cases}
      0 & \text{if iteration $t$ was drawn from the MDP}\\
      1 & \text{if iteration $t$ was a replay iteration}
    \end{cases}    
\end{align*}  
For notation simplicity, we write: $r_t$ instead of $r_t(s_t,a_t,s_{t+1})$.
We have that:
\begin{align*}
& Pr(s_{t+1} = s, r_{t} = r | s_t, a_t) \\
& = Pr(s_{t+1} = s, r_{t} = r | s_t, a_t, X_t = 1) \cdot Pr(X_t = 1 | s_t, a_t) \\
& + Pr(s_{t+1} = s, r_{t} = r | s_t, a_t, X_t = 0) \cdot Pr(X_t = 0 | s_t, a_t) \\
& = \frac{w_t(s_t,a_t,s,r)}{\sum_{\bar{s} \in S} \sum_{\bar{r} = r_{min}}^{r_{max}} w_t(s_t,a_t,\bar{s},\bar{r})} \cdot \frac{Pr(s_t,a_t|X_t = 1) \cdot Pr(X_t = 1)}{Pr(s_t,a_t|X_t = 1) \cdot Pr(X_t = 1) + Pr(s_t,a_t|X_t = 0) \cdot Pr(X_t = 0)} \\
& + \frac{q(s,r|s_t,a_t)}{\sum_{\bar{s} \in S} \sum_{\bar{r} = r_{min}}^{r_{max}} w_t(s_t,a_t,\bar{s},\bar{r})} \cdot \frac{Pr(s_t,a_t|X_t = 0) \cdot Pr(X_t = 0)}{Pr(s_t,a_t|X_t = 1) \cdot Pr(X_t = 1) + Pr(s_t,a_t|X_t = 0) \cdot Pr(X_t = 0)} \\
& = \frac{w_t(s_t,a_t,s,r)}{\sum_{\bar{s} \in S} \sum_{\bar{r} = r_{min}}^{r_{max}} w_t(s_t,a_t,\bar{s},\bar{r})} \cdot \frac{v \cdot \sum_{\bar{s} \in S} \sum_{\bar{r} = r_{min}}^{r_{max}} w_t(s_t,a_t,\bar{s},\bar{r})}{v \cdot \sum_{\bar{s} \in S} \sum_{\bar{r} = r_{min}}^{r_{max}} w_t(s_t,a_t,\bar{s},\bar{r}) + (1-v) \cdot \sum_{\bar{s} \in S} \sum_{\bar{r} = r_{min}}^{r_{max}} q(\bar{s},\bar{r}|s_t,a_t)} \\
& + \frac{q(s,r|s_t,a_t)}{\sum_{\bar{s} \in S} \sum_{\bar{r} = r_{min}}^{r_{max}} q(\bar{s},\bar{r}|s_t,a_t)} \cdot \frac{(1-v) \cdot \sum_{\bar{s} \in S} \sum_{\bar{r} = r_{min}}^{r_{max}} q(\bar{s},\bar{r}|s_t,a_t)}{v \cdot \sum_{\bar{s} \in S} \sum_{\bar{r} = r_{min}}^{r_{max}} w_t(s_t,a_t,\bar{s},\bar{r}) + (1-v) \cdot \sum_{\bar{s} \in S} \sum_{\bar{r} = r_{min}}^{r_{max}} q(\bar{s},\bar{r}|s_t,a_t)} \\
& = \frac{v \cdot w_t(s_t,a_t,s,r) + (1-v) \cdot q(s,r|s_t,a_t)}{\sum_{\bar{s} \in S} \sum_{\bar{r} = r_{min}}^{r_{max}} v \cdot w_t(s_t,a_t,\bar{s},\bar{r}) + (1-v) \cdot q(\bar{s},\bar{r}|s_t,a_t)} \\
& = (1-v) \cdot \phi_{1,t}(s_t,a_t,s,r) + v \cdot \phi_{2,t}(s_t,a_t,s,r)
\end{align*}

Thus, we have that:
\begin{align*}
& E[r_t(s_t,a_t, s_{t+1}) | F_t, s_{t+1}] = \sum_{r = r_{min}}^{r_{max}} r \cdot Pr(r_t = r | s_t, a_t, s_{t+1}) \\
& = \sum_{r = r_{min}}^{r_{max}} r \cdot \frac{Pr(r_t = r, s_{t+1} | s_t, a_t)}{Pr(s_{t+1}|s_t,a_t)} = \sum_{r = r_{min}}^{r_{max}} r \cdot \frac{(1-v) \phi_{1,t}(s_t,a_t,s_{t+1},r) + v \cdot \phi_{2,t}(s_t,a_t,s_{t+1},r)}{\sum_{\bar{r} = r_{min}}^{r_{max}} (1-v) \phi_{1,t}(s_t,a_t,s_{t+1},\bar{r}) + v \cdot \phi_{2,t}(s_t,a_t,s_{t+1},\bar{r})} \\
& = \tilde{R}_t(s_t,a_t,s_{t+1})
\end{align*}
and:
\begin{align*}
& Pr(s_{t+1} | F_t) = Pr(s_{t+1} | s_t,a_t) = \sum_{r = r_{min}}^{r_{max}} Pr(s_{t+1}, r | s_t,a_t) \\
& = \sum_{r = r_{min}}^{r_{max}} (1-v) \cdot \phi_{1,t}(s_t,a_t,s_{t+1},r) + v \cdot \phi_{2,t}(s_t,a_t,s_{t+1},r) \\
& = \tilde{p}_t(s'|s,a)
\end{align*}
Finally, this gives us the required condition on $e_t$, namely, that $E[e_t|F_t] = 0$. This is because we have:
\begin{align*}
& E[e_t(s_t,a_t)|F_t] \\
& = E[ r_t(s_t,a_t,s_{t+1})|F_t] + \gamma \cdot E[ max_{b \in A} Q_t(s_{t+1},b) |F_t] \\
&- \sum_{s' \in S} \tilde{p}_t(s'|s_t,a_t) \cdot \left( \tilde{R}_t(s_t,a_t,s') + \gamma \cdot max_{b \in A} Q_t(s',b) \right) \\
& = \left( \sum_{s' \in S} Pr(s_{t+1}=s'|F_t) \cdot E[ r_t(s_t,a_t,s_{t+1})|F_t, s_{t+1} = s'] \right) + \gamma \cdot E[ max_{b \in A} Q_t(s_{t+1},b) |F_t] \\
& - \sum_{s' \in S} \tilde{p}_t(s'|s_t,a_t) \cdot \left( \tilde{R}_t(s_t,a_t,s') + \gamma \cdot max_{b \in A} Q_t(s',b) \right) 
\end{align*}
\begin{align*}
& = \left( \sum_{s' \in S} \tilde{p}_t(s'|s_t,a_t) \cdot \tilde{R}_t(s_t,a_t,s_{t+1}) \right) + \gamma \cdot E[ max_{b \in A} Q_t(s_{t+1},b) |F_t] \\
& - \sum_{s' \in S} \tilde{p}_t(s'|s_t,a_t) \cdot \left( \tilde{R}_t(s_t,a_t,s') + \gamma \cdot max_{b \in A} Q_t(s',b) \right) \\
& = \gamma \cdot \sum_{s' \in S} Pr(s_{t+1}=s'|F_t) \cdot E[ max_{b \in A} Q_t(s_{t+1},b) |F_t, s_{t+1}=s'] - \gamma \cdot \sum_{s' \in S} \tilde{p}_t(s'|s_t,a_t)  \cdot max_{b \in A} Q_t(s',b) \\
& = \gamma \cdot \sum_{s' \in S} \tilde{p}_t(s'|s_t,a_t)  \cdot max_{b \in A} Q_t(s',b) - \sum_{s' \in S} \tilde{p}_t(s'|s_t,a_t)  \cdot max_{b \in A} Q_t(s',b) \\
& = 0
\end{align*} 
We now show that $E[e_t^2(s_t,a_t)|F_t] \leq K \cdot (1 + \| Q_t \|^2)$ . Since we have $\forall s\in S, a \in A, s' \in S: |r(s,a,s')| \leq \max(|r_{min}|, |r_{max}|)$, then $\tilde{R(s,a,s')} \leq \max(|r_{min}|, |r_{max}|)$. we can obtain for all $s,a,s',r$:
\begin{align*}
& E[e_t^2(s,a)|F_t] \\
& = E \left[\left( r(s,a,s') + \gamma \cdot max_{b \in A} Q_t(s',b) - \sum_{\bar{s} \in S} \tilde{p}_t(s,\bar{s}|a) \cdot \left( \tilde{R}_t(s,a,\bar{s}) + \gamma \cdot max_{b \in A} Q_t(\bar{s},b) \right)\right)^2 \bigg| F_t \right] \\
& = E \left[ \left( r(s,a,s') - \sum_{\bar{s} \in S} \tilde{p}_t(\bar{s}|s,a) \cdot \tilde{R}_t(s,a,\bar{s}) \right)^2 \bigg| F_t \right] \\
& + 2 \cdot \gamma \cdot E \left[ \left( r(s,a,s') - \sum_{\bar{s} \in S} \tilde{p}_t(\bar{s}|s,a) \cdot \tilde{R}_t(s,a,\bar{s}) \right) \cdot \left( max_{b \in A} Q_t(s',b) - \sum_{\bar{s} \in S} \tilde{p}_t(\bar{s}|s,a) \cdot max_{b \in A} Q_t(\bar{s},b) \right) \bigg| F_t \right] \\
& + \gamma^2 \cdot E \left[ \left( max_{b \in A} Q_t(s',b) - \sum_{\bar{s} \in S} \tilde{p}_t(\bar{s}|s,a) \cdot max_{b \in A} Q_t(\bar{s},b) \right)^2 \bigg| F_t \right] \\
& \leq 4 \cdot \max(|r_{min}|, |r_{max}|)^2 \\
& + 4 \cdot \gamma \cdot \max(|r_{min}|, |r_{max}|) \cdot E \left[  max_{b \in A} Q_t(s',b) - \sum_{\bar{s} \in S} \tilde{p}_t(\bar{s}|s,a) \cdot max_{b \in A} Q_t(\bar{s},b) \bigg| F_t \right] \\
& + \gamma^ 2 \cdot E \left[ max_{b \in A} Q_t^2 (s',b) \bigg| F_t \right] - 2 \gamma^ 2 \cdot E \left[ \sum_{\bar{s} \in S} \tilde{p}_t(\bar{s}|s,a) \cdot max_{b \in A} Q_t^2 (s',b) \cdot max_{b \in A} Q_t^2 (\bar{s},b) \bigg| F_t \right] \\
& + \gamma^2 \cdot E \left[ \left( \sum_{\bar{s} \in S} \tilde{p}_t(\bar{s}|s,a) \cdot max_{b \in A} Q_t(\bar{s},b) \right)^2 \bigg| F_t \right] \\
& \leq 4 \cdot \max(|r_{min}|, |r_{max}|)^2 \\
& + \gamma^ 2 \cdot E \left[ max_{s \in S, b \in A} Q_t^2 (s,b) \bigg| F_t \right] + \gamma^2 \cdot E \left[ \left( \sum_{\bar{s} \in S} \tilde{p}_t(\bar{s}|s,a) \cdot max_{s \in S, b \in A} Q_t(s,b) \right)^2 \bigg| F_t \right] \\
& \leq 4 \cdot \max(|r_{min}|, |r_{max}|)^2 + 2 \gamma^ 2 \cdot \| Q_t \|^2 
\end{align*}
Taking $K = 4 \cdot \max(|r_{min}|, |r_{max}|)^2 + 2 \gamma^2$, we obtain the required result.
Thus, algorithm~\ref{alg:Q-learning with replay} converges with probability 1.
\end{proof}

\subsection{Proof of Theorem~\ref{lemma:safe_policy_after_replay}}
\label{app:safe_policy_after_replay_proof}
\begin{proof}
Denote by $S_r(s_i,a_i) := \{s \in S : Pr(r(s_i,a_i,s) = r) > 0 \}$. \\
From assumption~\ref{assumption:replay_variance}, we have that if $s' \notin S_{r_{min}}(s_i,a_i)$, then:
\begin{align*}
\lim_{\frac{var(r(s_i,a_i))}{\sum_{s \in S, a \in A} var(r(s,a))} \rightarrow 1} w^\infty(s_i,a_i,s',r) = 0
\end{align*}
Otherwise, if $s' \in S_{r_{min}}(s_i,a_i)$, we have that:
\begin{align*}
\lim_{\frac{var(r(s_i,a_i))}{\sum_{s \in S, a \in A} var(r(s,a))} \rightarrow 1} w^\infty(s_i,a_i,s',r_{min}) = \frac{1}{|S_{r_{min}}(s_i,a_i)|}
\end{align*}
We look at the limit of $\phi_1^\infty, \phi_2^\infty$:
\begin{align*}
& \lim_{\frac{var(r(s_i,a_i))}{\sum_{s \in S, a \in A} var(r(s,a))} \rightarrow 1} \phi_1^\infty(s_i,a_i,s',r) \\ 
& = \lim_{\frac{var(r(s_i,a_i))}{\sum_{s \in S, a \in A} var(r(s,a))} \rightarrow 1} \frac{q(s',r|s_i,a_i)}{\sum_{\bar{s} \in S} \sum_{\bar{r} = r_{min}}^{r_{max}} (1-v) \cdot q(\bar{s},\bar{r}|s_i,a_i) + v \cdot w^\infty(s_i,a_i,\bar{s},\bar{r})} \\
& = \frac{q(s',r|s_i,a_i)}{\sum_{\bar{s} \in S} \sum_{\bar{r} = r_{min}}^{r_{max}} (1-v) \cdot q(\bar{s},\bar{r}|s_i,a_i) + v \cdot \frac{1}{|S_{{r_{min}}(s_i,a_i)}|} \cdot \mathbbm{1}_{\{\bar{r} = r_{min}\}}} \\
& = q(s',r|s_i,a_i)
\end{align*}
and:
\begin{align*}
& \lim_{\frac{var(r(s_i,a_i))}{\sum_{s \in S, a \in A} var(r(s,a))} \rightarrow 1} \phi_2^\infty(s_i,a_i,s',r) \\ 
& = \lim_{\frac{var(r(s_i,a_i))}{\sum_{s \in S, a \in A} var(r(s,a))} \rightarrow 1} \frac{w^\infty(s_i,a_i,s',r)}{\sum_{\bar{s} \in S} \sum_{\bar{r} = r_{min}}^{r_{max}} (1-v) \cdot q(\bar{s},\bar{r}|s_i,a_i) + v \cdot w^\infty(s_i,a_i,\bar{s},\bar{r})} \\
& = \frac{ \frac{1}{|S_{r_{min}}(s,a)|} \cdot \mathbbm{1}_{\{r = r_{min}\}}}{\sum_{\bar{s} \in S} \sum_{\bar{r} = r_{min}}^{r_{max}} (1-v) \cdot q(\bar{s},\bar{r}|s_i,a_i) + v \cdot  \frac{1}{|S_{r_{min}}(s_i,a_i)|} \cdot \mathbbm{1}_{\{\bar{r} = r_{min}\}}} \\
& = \frac{1}{|S_{r_{min}}(s_i,a_i)|} \cdot \mathbbm{1}_{\{r = r_{min}\}}
\end{align*}
Note that if $s' \notin S_r(s_i,a_i)$ then $\phi_2^\infty(s_i,a_i,s',r) = 0$.
This leads to the following - for $s' \in S_r(s_i,a_i)$:
\begin{align*}
& \lim_{\frac{var(r(s_i,a_i))}{\sum_{s \in S, a \in A} var(r(s,a))} \rightarrow 1} \tilde{p}^\infty (s_i,s'|a_i) = \sum_{r = r_{min}}^{r_{max}} (1-v) \cdot  q(s',r|s_i,a_i) + v \cdot \frac{1}{|S_{r_{min}}(s_i,a_i)|} \cdot \mathbbm{1}_{\{r = r_{min}\}} \\
& = (1-v) \cdot \left( \sum_{r = r_{min}}^{r_{max}} q(s',r|s_i,a_i) \right) + v \cdot \frac{1}{|S_{r_{min}}(s_i,a_i)|} 
\end{align*}
In the above expression, we see that the transition probabilities are shifted to having higher probability to transitions from state $s_i$ with action $a_i$ to states in which receiving a reward of $r_{min}$ is possible.
For $s' \notin S_r(s_i,a_i)$, we get:
\begin{align*}
& \lim_{\frac{var(r(s_i,a_i))}{\sum_{s \in S, a \in A} var(r(s,a))} \rightarrow 1} \tilde{p}^\infty (s'|s_i,a_i) = \sum_{r = r_{min}}^{r_{max}} (1-v) \cdot  q(s',r|s_i,a_i)
\end{align*}
We now inspect the effect of the replay scheme on the expected reward. For $s' \in S_r(s_i,a_i)$ we have:
\begin{align*}
& \lim_{\frac{var(r(s_i,a_i))}{\sum_{s \in S, a \in A} var(r(s,a))} \rightarrow 1} \tilde{p}^\infty(s'|s_i,a_i) \cdot \tilde{R}^\infty(s_i,a_i,s') \\
& = \lim_{\frac{var(r(s_i,a_i))}{\sum_{s \in S, a \in A} var(r(s,a))} \rightarrow 1} \sum_{r = r_{min}}^{r_{max}} r \cdot (1-v) \cdot \phi_1^\infty(s,a,s',r) + v \cdot \phi_2^\infty(s,a,s',r) \\
& = (1-v) \cdot \left( \sum_{r = r_{min}}^{r_{max}} r \cdot q(s',r|s_i,a_i) \right) + v \cdot r_{min} \cdot \frac{1}{|S_{r_{min}}(s_i,a_i)|} 
\end{align*}
The above shows that the expected reward is shifted towards $r_{min}$, and thus we have that the Q-value for $s_i,a_i$ will decrease as we do more replay.
For $s' \notin S_r(s_i,a_i)$ we get:
\begin{align*}
& \lim_{\frac{var(r(s_i,a_i))}{\sum_{s \in S, a \in A} var(r(s,a))} \rightarrow 1} \tilde{p}^\infty(s'|s_i,a_i) \cdot \tilde{R}^\infty(s_i,a_i,s') = \lim_{\frac{var(r(s_i,a_i))}{\sum_{s \in S, a \in A} var(r(s,a))} \rightarrow 1} \sum_{r = r_{min}}^{r_{max}} r \cdot (1-v) \cdot \phi_1^\infty(s,a,s',r) \\
& = (1-v) \cdot \left( \sum_{r = r_{min}}^{r_{max}} r \cdot q(s',r|s_i,a_i) \right)
\end{align*}
The Q-values we get are as follows:
\begin{align*}
& Q^\ast(s_i,a_i) = \sum_{s' \in S} \tilde{p}^\infty(s'|s_i,a_i) \cdot \left( \tilde{R}^\infty(s_i,a_i,s') + \gamma \cdot max_{b \in A} Q(s',b) \right) \\
& = \sum_{s' \in S_{r_{min}}} \tilde{p}^\infty(s'|s_i,a_i) \cdot \left( \tilde{R}^\infty(s_i,a_i,s') + \gamma \cdot max_{b \in A} Q(s',b) \right) \\
& + \sum_{s' \notin S_{r_{min}}} \tilde{p}^\infty(s'|s_i,a_i) \cdot \left( \tilde{R}^\infty(s_i,a_i,s') + \gamma \cdot max_{b \in A} Q(s',b) \right) \\
& = \sum_{s' \in S_{r_{min}}} (1-v) \cdot \left( \sum_{r = r_{min}}^{r_{max}} r \cdot q(s',r|s_i,a_i) \right) + v \cdot r_{min} \cdot \frac{1}{|S_{r_{min}}(s_i,a_i)|} \\
& + \sum_{s' \in S_{r_{min}}} \gamma \cdot \left( (1-v) \cdot \left( \sum_{r = r_{min}}^{r_{max}} q(s',r|s_i,a_i) \right) + v \cdot \frac{1}{|S_{r_{min}}(s_i,a_i)|} \right) \cdot \max_{b \in A} Q^\ast(s',b) \\
& + \sum_{s' \notin S_{r_{min}}} (1-v) \cdot \left( \sum_{r = r_{min}}^{r_{max}} r \cdot q(s',r|s_i,a_i) \right) \\
& + \sum_{s' \notin S_{r_{min}}} \gamma \cdot \left(  \sum_{r = r_{min}}^{r_{max}} (1-v) \cdot  q(s',r|s_i,a_i) \right) \cdot \max_{b \in A} Q^\ast(s',b) \\
& = (1-v) \cdot \left( \sum_{s' \in S} \left( \sum_{r = r_{min}}^{r_{max}} r \cdot q(s',r|s_i,a_i) \right) + \gamma \cdot  \left(  \sum_{r = r_{min}}^{r_{max}} (1-v) \cdot  q(s',r|s_i,a_i) \right) \cdot \max_{b \in A} Q^\ast(s',b) \right) \\
& + v \cdot \sum_{s' \in S_{r_{min}}} r_{min} \cdot \frac{1}{|S_{r_{min}}(s_i,a_i)|} + \gamma \cdot \frac{1}{|S_{r_{min}}(s_i,a_i)|} \cdot \max_{b \in A} Q^\ast(s',b) \\
& = (1-v) \cdot \left( R(s_i,a_i) + \gamma \cdot \sum_{s' \in S} p(s'|s_i,a_i) \cdot \max_{b \in A} Q^\ast(s',b) \right) \\
& + v \cdot \left( r_{min} + \gamma \cdot \sum_{s' \in S_{r_{min}}} \frac{1}{|S_{r_{min}}(s_i,a_i)|} \cdot \max_{b \in A} Q^\ast(s',b) \right)
\end{align*}
We compare the above Q-values to the Q-value that would have been received in the same MDP but where no replay was performed, i.e. if $v=0$. Denote the Q-value in this case by $Q_{v=0}^\ast$. 
In such a case we get:
\begin{align*}
Q_{v=0}^\ast(s_i,a_i) = R(s_i,a_i) + \gamma \cdot \sum_{s' \in S} p(s'|s_i,a_i) \cdot \max_{b \in A} Q_{v=0}^\ast(s',b)
\end{align*}
comparing this to the Q-value we get:
\begin{align*}
& Q^\ast(s_i,a_i) = (1-v) \cdot \left( R(s_i,a_i) + \gamma \cdot \sum_{s' \in S} p(s'|s_i,a_i) \cdot \max_{b \in A} Q^\ast(s',b) \right) \\
&+ v \cdot \left( r_{min} + \gamma \cdot \sum_{s' \in S_{r_{min}}} \frac{1}{|S_{r_{min}}(s_i,a_i)|} \cdot \max_{b \in A} Q^\ast(s',b) \right)
\end{align*}
We see that the resulting Q-values are lower due to the up-sampling of transitions which result in next-states and rewards that are low.
Now assume that $\forall b\in A: Q_{v=0}^\ast(s_i,a_i) \geq Q_{v=0}^\ast(s_i,b)$, i.e. the optimal action in $s_i$ when no replay is done was the action $a_i$.
Since $\forall b \in A: w^\infty(s_i,b,\cdot,\cdot) \rightarrow 0$ as $\frac{var(s_i,a_i)}{\sum_{s \in S}\sum {a \in A} var(r(s,a))} \rightarrow 1$, we get that for all other state-action pairs, $Q^\ast(s_i,b) \rightarrow Q_{v=0}^\ast(s_i,b)$, thus, if:
\begin{align*}
& \exists b \in A : Q_{v=0}^\ast(s_i,a_i) - Q_{v=0}^\ast(s_i,b) < Q_{v=0}^\ast(s_i,a_i) - Q^\ast(s_i,a_i) \\
&\leq v \cdot \left(  R(s_i,a_i) - r_{min} \right) \\
& + \gamma \cdot \sum_{s' \in S} p(s'|s_i,a_i) \cdot \left( \max_{b \in A} Q_{v=0}^\ast(s',b) - \max_{b \in A} Q^\ast(s',b)\right) \\
& + v \cdot \gamma \cdot \left( \sum_{s' \in S}  p(s'|s_i,a_i)  \cdot \max_{b \in A} Q^\ast(s',b) - \sum_{s' \in S_{r_{min}}} \frac{1}{|S_{r_{min}}(s_i,a_i)|} \cdot \max_{b \in A} Q^\ast(s',b) \right)
\end{align*}
Then, $Q^\ast(s_i,b) > Q^\ast(s_i,a_i)$ and thus the policy has changed to choosing the less variable action.
\end{proof}

\subsection{Proof of Lemma~\ref{lemma:variance_based_replay_convergence}}
\label{app:variance_based_replay_convergence_proof}
\begin{proof}
Denote by $I_t(s,a)$ the set of iterations in which a transition from state $s$ with action $a$ occurred, and $|I_t(s,a)|$ its size. Denote by $\bar{r} = \frac{\sum_{i \in I_t(s,a)} r_i}{|I_t(s,a)|}$, the empirical mean of the reward of transitions from $s$ with action $a$. \\
We use the GLIE assumption, namely, that every state and action are visited infinitely often. Thus, $\lim_{t \rightarrow \infty} |I_t(s,a)| = \infty$. \\
This means that for each state and action $s,a$, we have that:
\begin{align*}
& var_t(r(s,a)) = \frac{\sum_{i \in I_t(s,a)} (r_i - \bar{r})^2}{|I_t(s,a)|} = \frac{\sum_{i \in I_t(s,a)} r_i^2}{|I_t(s,a)|} - \left( \frac{\sum_{i \in I_t(s,a)} r_i}{|I_t(s,a)|} \right)^2
\end{align*}
Since $r_i$ are i.i.d and bounded (and thus have a finite mean), we have from the strong law of large numbers that: $\lim_{t \rightarrow \infty} \frac{\sum_{i \in I_t(s,a)} (r_i - \bar{r})^2}{|I_t(s,a)|} = \frac{\sum_{i \in I_t(s,a)} r_i^2}{|I_t(s,a)|} \rightarrow E[r_i^2]$.
In addition, again by the strong law of large number we have: $\lim_{t \rightarrow \infty} \frac{\sum_{i \in I_t(s,a)} r_i}{|I_t(s,a)|} \rightarrow E[r_i]$, and thus: $\lim_{t \rightarrow \infty} \left( \frac{\sum_{i \in I_t(s,a)} r_i}{|I_t(s,a)|} \right)^2 \rightarrow E[r_i]^2$. We conclude that: $\lim_{t \rightarrow \infty} var_t(r(s,a)) = E[r_i^2] - E[r_i]^2 = var(r(s,a))$. \\
In addition, we have that as $t \rightarrow \infty$, then: 
\begin{align*}
& \forall r \in [r_{min}, r_{max}]: \lim_{t \rightarrow \infty} Pr(r \notin M_t[r(s,a)]) = \lim_{t \rightarrow \infty} \prod_{i \in I_t(s,a) } Pr(r_i \neq r) \rightarrow 0
\end{align*}
and:
\begin{align*}
& \forall s \in S, a \in A, s' \in S_r(s,a): \lim_{t \rightarrow \infty} Pr(s' \notin L_t[s,a,r]) = \lim_{t \rightarrow \infty} \prod_{i \in I_t(s,a) } Pr(s_{i+1} \neq s) \rightarrow 0
\end{align*}
This stems from the GLIE assumption, that lead to: $\lim_{t \rightarrow 0} |I_t(s,a)| = \infty$, and since we have a product of probabilities (smaller than $1$) with $|I_t(s,a)| \rightarrow \infty$, we get this limit. Thus, we conclude that $\lim_{t \rightarrow \infty M_t(r(s,a))} = [r_{min}, r_{max}]$ and $\lim_{t \rightarrow \infty L_t(s,a,r) = S_r(s,a)}$.
Thus, we have that: 
\begin{align*}
&\lim_{t \rightarrow \infty} w_t(s,a,s',r) = \begin{cases}
 \frac{var(r(s,a))}{\sum_{\bar{s} \in S} \sum_{\bar{a} \in A} var(r(\bar{s},\bar{a})))} \cdot \frac{e^{-\beta  r}}{\sum_{r' = r_{min}}^{r_{max}} e^{-\beta  r'}} \cdot \frac{1}{|S_r(s,a)|} & \text{if $s' \in S_r(s,a)$} \\
0 & \text{o.w.}
\end{cases}
\end{align*}
\end{proof}

\end{document}